\documentclass[10pt,twocolumn,letterpaper]{article}

\usepackage[pagenumbers]{wacv} 

\usepackage{multirow} 

\definecolor{wacvblue}{rgb}{0.21,0.49,0.74}
\usepackage[pagebackref,breaklinks,colorlinks,allcolors=wacvblue]{hyperref}

\def\wacvPaperID{3322} 
\def\confName{WACV}
\def\confYear{2026}

\title{With Great Context Comes Great Prediction Power: \\ Classifying Objects via Geo-Semantic Scene Graphs}

\author{Ciprian Constantinescu \; \; \; \; \; \; \; \; \; \; \; Marius Leordeanu\\ \\
National University of Science and Technology POLITEHNICA Bucharest \\
Splaiul Independenței 313, Bucharest 060042 \\
{\tt\small leordeanu@gmail.com}
}

\begin{document}
\maketitle
\begin{abstract}
Humans effortlessly identify objects by leveraging a rich understanding of the surrounding scene, including spatial relationships, material properties, and the co-occurrence of other objects. In contrast, most computational object recognition systems operate on isolated image regions, devoid of meaning in isolation, thus ignoring this vital contextual information. This paper argues for the critical role of context and introduces a novel framework for contextual object classification. We first construct a Geo-Semantic Contextual Graph (GSCG) from a single monocular image. This rich, structured representation is built by integrating a metric depth estimator with a unified panoptic and material segmentation model. The GSCG encodes objects as nodes with detailed geometric, chromatic, and material attributes, and their spatial relationships as edges. This explicit graph structure makes the model's reasoning process inherently interpretable. We then propose a specialized graph-based classifier that aggregates features from a target object, its immediate neighbors, and the global scene context to predict its class. Through extensive ablation studies, we demonstrate that our context-aware model achieves a classification accuracy of 73.4\%, dramatically outperforming context-agnostic versions (as low as 38.4\%). Furthermore, our GSCG-based approach significantly surpasses strong baselines, including fine-tuned ResNet models (max 53.5\%) and a state-of-the-art multimodal Large Language Model (LLM), Llama 4 Scout, which, even when given the full image alongside a detailed description of objects, maxes out at 42.3\%. These results on COCO 2017 train/val splits highlight the superiority of explicitly structured and interpretable context for object recognition tasks.
\end{abstract}
    
\section{Introduction}

The task of object recognition is a cornerstone of computer vision. Traditional approaches have achieved remarkable success by training deep neural networks to classify objects based on their intrinsic visual features \cite{krizhevsky2012imagenet, he2016deep, simonyan2014very, szegedy2015going}. However, these models often operate in a contextual vacuum, treating each object as an independent entity. This paradigm starkly contrasts with human perception. For example, when identifying a blurry object on a wooden surface next to a keyboard, a person naturally infers it is likely a "mouse," using the surrounding context to resolve ambiguity.

We argue that for artificial intelligence to achieve human-level understanding, it must similarly learn to utilize the power of context. The identity of an object is deeply intertwined with its environment: its physical material, its size and shape, what other objects it is near or touching, and the overall semantics of the scene \cite{HoughNet, zellers2018neural}.

In this work, we propose a complete pipeline to operationalize this idea. We introduce a method to transform the dense, unstructured pixels of a single 2D image into a high-level, queryable, and interpretable Geo-Semantic Contextual Graph (GSCG). This graph serves as a structured world model, abstracting away from the pixel-level to a representation of objects, their rich properties, and their interrelations.

Our contributions are threefold:
\begin{enumerate}
    \item We present a novel method for constructing a GSCG from a monocular image by fusing outputs from a metric depth predictor with those of a segmentation model fine-tuned on a merged ontology of panoptic and material datasets to jointly predict instance, semantic class, and material labels.
    \item We develop a graph-based classifier that explicitly leverages the GSCG's structure. It reasons about a target object by aggregating its intrinsic features (geometry, material, color) with the features of its neighbors and the global semantic context of the scene. This approach is inspired by relational reasoning networks \cite{santoro2017simple, battaglia2018relational} and graph neural networks \cite{kipf2016semi, velivckovic2017graph, hamilton2017inductive}.
    \item Through comprehensive experiments on a contextual object classification task using the COCO 2017 train/val splits \cite{lin2014microsoft}, we demonstrate the immense value of context. Our full model (73.4\% accuracy) vastly outperforms ablated versions lacking specific contextual cues. Moreover, it establishes a new state-of-the-art compared to powerful baselines like fine-tuned Convolutional Neural Networks (CNNs) (53.5\%) and a state-of-the-art multimodal LLM (42.3\%), proving that structured, explicit, and interpretable context representation is more effective than implicit context learning in raw pixels or unstructured text.
\end{enumerate}

This work paves the way for more robust and intelligent vision systems, with future applications in scene understanding, robotics, and even the contextual inference of novel object classes.

\section{Related Work}

Our research builds upon several key areas in computer vision and AI.

\paragraph{Monocular 3D Scene Understanding.}
Inferring 3D structure from a single image is a classic, ill-posed problem \cite{bar1964some}. Recent advancements in deep learning have produced powerful models for monocular depth estimation \cite{Bochkovskii2024:arxiv, li2018factorizable}. Concurrently, panoptic segmentation \cite{kirillov2019panoptic} unifies semantic and instance segmentation to provide a comprehensive 2D parse of a scene. Models like OneFormer \cite{jain2023oneformer} use a universal transformer architecture for various segmentation tasks. Our work integrates these streams, but goes a step further by also incorporating material properties into a unified 3D representation, which we then structure into a graph.

\paragraph{Scene Graphs.}
The concept of a scene graph, a structured data representation of objects and their relationships, has its roots in 3D computer graphics from the 1980s \cite{schuster2015generating}. Only more recently has it been widely adopted for computer vision tasks. Initial works on visual scene graphs \cite{johnson2015image, lu2016visual} have emerged as a powerful structured representation, capturing objects as nodes and their relationships as edges. While primarily used in 2D, extensions to 3D have been proposed \cite{armeni20193d, wald2020learning}. However, these graphs are often limited to semantic labels and bounding boxes. Our GSCG provides a much richer set of attributes, including fine-grained material composition, dominant colors per material, and metric 3D geometry (size, orientation), all derived from a single image.

\paragraph{Contextual Object Recognition.}
The idea that context aids recognition is well-established \cite{winograd1972understanding}. Early work used statistical priors of object co-occurrence \cite{misra2015seeing, galleguillos2010context}. More recently, context has been implicitly captured by large receptive fields in CNNs \cite{he2016deep} or through attention mechanisms in transformers \cite{vaswani2017attention, dosovitskiy2020image}. Other approaches explicitly model context using graphical models \cite{chen2019knowledge, yang2018graph} or by reasoning about object interactions \cite{woo2022tackling}. Our approach differs by making context explicit and structured. Instead of asking the model to learn context implicitly from pixels, we provide it directly through the GSCG, allowing for more direct and interpretable reasoning.

\section{Methodology}
Our methodology is divided into two main stages: first, the generation of the Geo-Semantic Contextual Graph (GSCG) from a monocular image; and second, the use of this graph in a specialized classifier to predict an object's class based on its context.

\subsection{Geo-Semantic Contextual Graph (GSCG) Generation}

\begin{figure*}[t]
\centering
\includegraphics[width=0.9\textwidth]{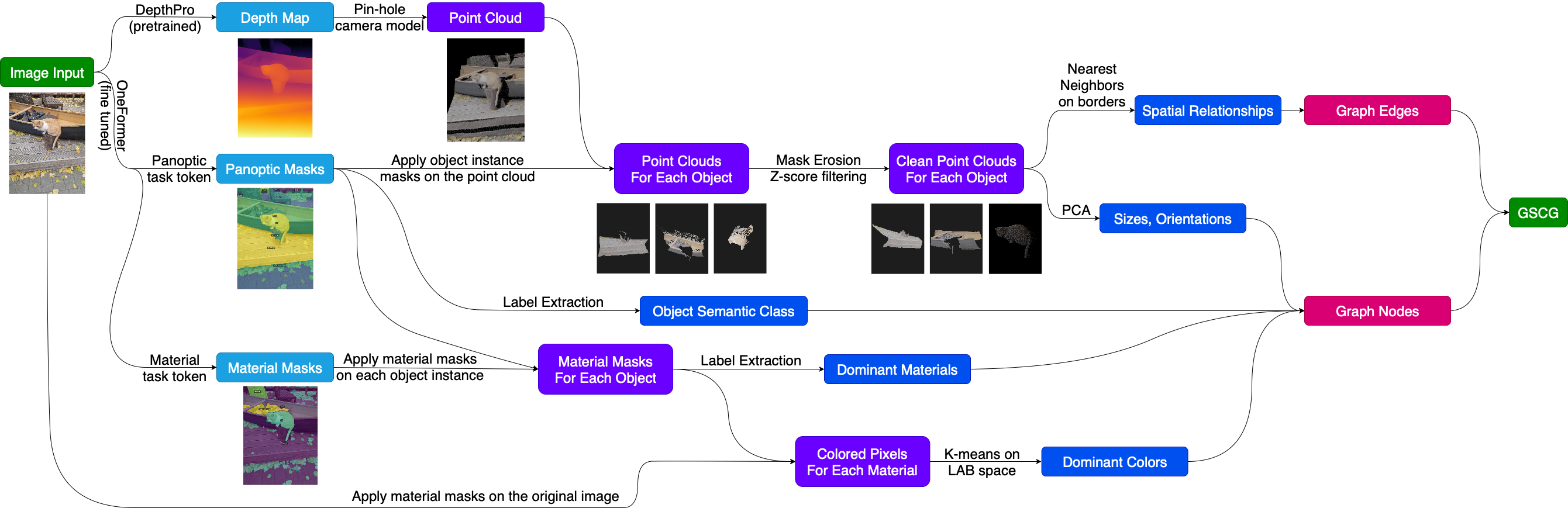}
\caption{The Geo-Semantic Contextual Graph (GSCG) Generation Pipeline. From a single RGB image, we (a) estimate metric depth and perform unified segmentation to get panoptic and material masks. (b) These are fused to create per-instance 3D point clouds. (c) Rich attributes (geometry, material, color) are extracted for each instance. (d) This information is encoded into a GSCG, where nodes represent objects and edges represent spatial relationships. This structured representation then feeds our classifier.}
\label{fig:pipeline}
\end{figure*}

The GSCG is the cornerstone of our approach. Its construction is a procedural pipeline designed to abstract a rich, structured world model from raw pixels. The full pipeline is illustrated in Figure \ref{fig:pipeline}.

\subsubsection{Unified Scene Perception}
To capture a holistic understanding of the scene, we require geometry, instances, semantic categories, and material properties. \\
\begin{itemize}
    \item \textbf{Geometric Foundation:} We use a pre-trained model, DepthPro \cite{Bochkovskii2024:arxiv}, to estimate a metric depth map from the input RGB image. The model also infers the camera's focal length, allowing us to back-project each pixel into a 3D point using a pinhole camera model, thus forming a dense point cloud of the entire scene.
    \item \textbf{Unified Segmentation:} To achieve both panoptic and material segmentation from a single model, we fine-tuned a OneFormer transformer \cite{jain2023oneformer} on a unified ontology. To achieve this, we combined the COCO Panoptic dataset (using the high-quality COCOnut subset \cite{coconut2024cvpr}) and the Dense Material Segmentation (DMS) dataset \cite{dmsdataset}, using only their official training splits to prevent any data leakage into our classifier's validation set. A unified set of 175 labels was created, encompassing both object categories from COCO and material types from DMS. During training and inference, the model is conditioned with a task prompt ($panoptic$ or $material$), enabling it to output instance IDs, semantic labels, and material labels from the same underlying representation. The OneFormer model is trained for 30 epochs using the AdamW optimizer \cite{loshchilov2017decoupled} (lr=1e-5) with a cosine annealing warm restarts scheduler, batch size 6, and input resolution of 800 by 800. This joint training encourages the model to learn intrinsic correlations between objects and their typical materials (e.g., that $cars$ are often made of $metal$ and $glass$).
\end{itemize}

\subsubsection{Procedural Graph Construction}
With depth and unified segmentation maps, we build the graph procedurally: \\
\begin{enumerate}
    \item \textbf{Instance Point Cloud Creation:} For each object instance identified by the panoptic segmentation, we gather all corresponding 3D points from the dense scene cloud. To mitigate artifacts from depth estimation, especially at object boundaries, we first apply a 2D erosion to the instance mask (kernel size 3x3) and then a statistical outlier removal filter. This filter removes points whose per-axis Z-score is above a threshold of 2.5, effectively cleaning the point cloud. Instances with fewer than 100 valid 3D points after filtering are discarded. This point cloud processing is inspired by techniques used in PointNet \cite{qi2017pointnet, qi2017pointnet++}.
    \item \textbf{Node Attribute Extraction:} Each cleaned instance point cloud becomes a node in our graph. We extract a rich set of attributes for each node:
    \begin{itemize}
        \item \textit{Geometric}: We apply Principal Component Analysis (PCA) to the point cloud coordinates. The eigenvalues provide an estimate of the object's 3D size along its principal axes, and the eigenvectors define its orientation. The centroid is the mean of the coordinates.
        \item \textit{Material (Textural)}: We overlay the instance's 2D mask onto the material segmentation map to determine its composition (e.g., 70\% wood, 30\% metal), keeping the top 3 materials by area.
        \item \textit{Chromatic}: For each material on an object, we extract its dominant colors. We perform k-means clustering (k=20, $k$-$means{++}$ initialization, seed 42) on the LAB color space of the corresponding pixels. We then merge clusters with a CIEDE2000 distance below 15.0. The final colors are named by finding the nearest color in the CSS4 palette. This yields a list of named colors with percentages (e.g., 80\% $light$-$gray$, 20\% $dark$-$red$).
        \item \textit{Semantic}: The object's class label (e.g., 'table') is taken from the panoptic segmentation.
    \end{itemize}
    \item \textbf{Edge Creation:} We define edges between nodes to represent spatial relationships. A brute-force nearest-neighbor search is used to determine proximity between the point clouds of every pair of objects.
    Two objects are considered $touch$ing if a significant portion of their points located at their boundaries are in close proximity. Specifically, for a pair of objects (A, B), we calculate the percentage of points in B that are within a 5 cm radius of any point in A. If this percentage exceeds a 5\% threshold, a $touch$ edge is created with the percentage value as its weight.
    Two objects are $near$ if their centroids are within 1.0 m. The weight for a $near$ edge is calculated as an exponential function of the inverse normalized distance, specifically \(e^{-d/\bar{d}}\), where \(d\) is the centroid distance and \(\bar{d}\) is the mean centroid distance across all object pairs in the scene.
    If a $touch$ edge already exists between two objects that are also $near$, the edge type is updated to $mixed$, and the calculated $near$ weight is added to the existing $touch$ weight.
\end{enumerate}
The final output is a Geo-Semantic Contextual Graph (GSCG), a powerful, interpretable, and queryable representation of the scene's content and structure.

\subsection{Contextual Object Classification}
To demonstrate the GSCG's predictive power, we design a classification task: given a graph representing a scene where one target object's semantic label is hidden, predict that label. We develop a $GraphObjectClassifier$ to solve this task. Its architecture is specifically designed to process the GSCG's flexible structure, where each node can have a variable number of neighbors and of extracted features (textural and chromatic).

The model takes the GSCG and a target node ID as input. Its reasoning process unfolds in four steps:

First, \textbf{Feature Embedding}, where dedicated Multi-Layer Perceptrons (MLPs) process the different features of each object. The model uses a modular design with separate embedders for geometric, material, and color features. For instance, the $ColorEmbedder$ uses a small MLP to transform CIELAB color values into a latent space, which are then weighted by their percentage. The $MaterialEmbedder$ combines the embedding of the material's semantic label with the embedding of its dominant colors. Finally, the $ObjectEmbedder$ concatenates the embeddings from the object's label (if available), its geometry, and its materials. This design allows for a flexible combination of features. Crucially, the target object's semantic label is omitted from its embedding.

Second, \textbf{Local Context Aggregation}, where the model focuses on the immediate neighborhood. The embeddings of the target's direct neighbors are concatenated with embeddings for the edge type and weight. We use a graph attention mechanism \cite{velivckovic2017graph} to weigh the importance of each neighbor based on its features and the edge type connecting it to the target. This allows the model to dynamically focus on the most relevant local contextual cues.

Third, \textbf{Global Context Integration}, where the model considers the broader scene. A global context vector is created by generating a histogram of the semantic labels of all other objects in the scene. This raw count vector is passed through an MLP, capturing the overall scene type (e.g., a scene with a bed, wardrobe, and dresser is likely a "bedroom").

Fourth, \textbf{Classification}, where the embeddings of the target object, its aggregated local context, and the global scene context are concatenated and passed through a final MLP classifier to predict the target object's class. This final step is similar to the readout phase in message passing neural networks \cite{gilmer2017neural}.

\section{Experiments}
We conduct a series of experiments to validate our central hypothesis: that explicit, structured context is crucial for robust object recognition.

\subsection{Experimental Setup}
\paragraph{Dataset and Task.} We use the COCO 2017 train/val splits \cite{lin2014microsoft} containing 118k training images and 5k validation images with 80 object categories. For each image, we generate its GSCG. We then create our classification task: for each object instance in the graph, we mask its true label and ask the model to predict it using only the available context. The context available depends on the specific model being tested.

\paragraph{Evaluation Metric.} We report top-1 classification accuracy on the validation split of COCO 2017 \cite{lin2014microsoft}. For our GSCG classifier variants and the Llama 4 Scout and ResNet baselines, we compute bootstrap confidence intervals using 1,000 bootstrap samples. The resampling unit for this analysis is the object instance, to assess the stability of the model's predictions across individual objects.

\paragraph{Our Models.} We test several versions of our $GraphObjectClassifier$ to perform an extensive ablation study. The $full\_model$ uses all available GSCG features. Other versions systematically remove features: $no\_geometry$, $no\_materials$, $no\_neighbors$ (removes local, precise context), $no\_extended\_context$ (removes global, superficial context), and combinations thereof. The $minimal\_model$ ($no\_materials\_no\_extended\_context\_no\_geometry$) represents a severely limited version that can only predict an object's class from the semantic labels of its immediate neighbors. For models without neighbors ($no\_neighbors$), we use a simplified $NeighborlessObjectClassifier$ architecture that processes only the target object's intrinsic features and global scene context.

\paragraph{Training Details.} All GSCG classifier variants are trained on the COCO 2017 training set for 20 epochs using the AdamW optimizer \cite{loshchilov2017decoupled} with a learning rate of 1e-4, batch size of 512, and weight decay of 1e-4. We use early stopping based on validation F1-score with a patience of 10 epochs, saving the best model based on the validation F1-score. The ResNet baselines are similarly trained on the COCO 2017 training set using a ResNet-101 architecture pre-trained on ImageNet, trained for 200 epochs with the AdamW optimizer (lr=1e-4), batch size 512 for context, object and combines modes, using standard data augmentation (random horizontal flips, rotations up to 15°, and color jittering). All models are implemented in PyTorch and trained on NVIDIA A100 GPUs using Hugging Face Accelerate for distributed training.

\subsection{Baselines}
We compare our method against two strong baselines designed to test different ways of leveraging context. \\
\begin{enumerate}
    \item \textbf{Fine-tuned ResNet \cite{he2016deep}:} A powerful CNN baseline using a ResNet-101 architecture pre-trained on ImageNet. We test three variants on an "image inpainting" version of our task:
    \begin{itemize}
        \item \textit{Object-Only:} The model sees only a cropped image of the target object against a black background. This tests classification based on intrinsic appearance alone.
        \item \textit{Context-Only:} The model sees the full image, but the target object is blacked out. This tests the ability to infer an object from its visual surroundings.
        \item \textit{Combined:} The model receives a 6-channel input: the 3-channel RGB image where the target object is blacked out and the background visible plus the 3-channel image where the target object is isolated and the background is black. This provides both the object and its context.
    \end{itemize}
    \item \textbf{Multimodal Large Language Model (LLM):} We use Llama 4 Scout \cite{meta2025llama4scout}, a state-of-the-art multimodal model. We evaluate it in several settings to mirror our ResNet baselines and test its reasoning capabilities on unstructured data:
    \begin{itemize}
        \item \textit{Text-Only, Zero-Shot}: We generate a detailed textual description of the GSCG context and prompt the LLM to guess the object.
        \item \textit{Text-Only, Few-Shot}: In addition to the textual description, we provide an example of GSCG-powered contextual description for each of the 10 most abundant object categories in the validation dataset, alongside their corresponding class labels.
        \item \textit{Multimodal, Zero-Shot}: We provide both the generated description and the original image, with the target object visible, thus facilitating even more context for the LLM to reason about the object.
        \item \textit{Image-Only, Context}: We provide only the image with the target object blacked out (matching the Context-Only ResNet task) and prompt the LLM to predict the missing object's semantic class without any textual description derived from the GSCG.
    \end{itemize}
\end{enumerate}

\subsection{Results and Discussion}

\paragraph{The Power of Structured Context.}
Table \ref{tab:main_results} shows a clear performance hierarchy. Our $full\_model$ achieves an accuracy of 73.43 ± 0.32\%, establishing a new state-of-the-art on this task. It surpasses the best baselines —the combined ResNet (53.52\%) and the multimodal Llama 4 Scout (42.34\%)— by a significant margin. This difference underscores the effectiveness of representing scenes as explicit, structured graphs. While the ResNet and LLM models see all the pixels, they struggle to disentangle and reason about the complex interplay of geometric, material, and spatial relationships that our GSCG model handles explicitly. The fact that our model outperforms a next-generation multimodal LLM, even when the LLM is given the full image, is a powerful testament to our approach.

\paragraph{Ablation Study: Dissecting Context.}
Our GSCG classifier's ablation study, detailed in the last column of Table \ref{tab:llm_comprehensive}, quantifies the importance of each contextual component.
\begin{itemize}
    \item \textbf{All Context is Better:} The $full\_model$ (73.43 ± 0.32\%) is the undisputed champion, showing that integrating all available information is optimal.
    \item \textbf{Spatial Context is Key:} Removing neighbors ($no\_neighbors$, 71.33 ± 0.33\%) results in a moderate decrease in accuracy, while removing the global scene context ($no\_extended\_context$, 68.55 ± 0.34\%) produces a more substantial performance degradation. Removing both components ($no\_neighbors\_no\_extended\_context$, 56.66 ± 0.37\%) severely impacts performance, reducing accuracy by over 16 points. This demonstrates that spatial relationships and scene-level context are critical factors for accurate object recognition.
    \item \textbf{Physical Properties Matter:} Removing an object's geometric properties ($no\_geometry$, 67.70 ± 0.33\%) or its material composition ($no\_materials$, 61.84 ± 0.35\%) results in a considerable performance drop. This validates our effort to extract these rich physical attributes. An object's shape and substance are core to its identity. The $minimal\_model$, with no geometry, material or spatial context, achieves 38.41 ± 0.36\% accuracy, barely better than the worst baseline.
\end{itemize}

\paragraph{LLM Performance Analysis.}
The detailed LLM results in Table \ref{tab:llm_comprehensive} are particularly insightful.
First, providing the image alongside the text ($Multimodal$) gives a massive performance boost over text-only prompting, jumping from a max of 28.40\% to 42.34\%. This shows the LLM is heavily relying on the visual data to resolve ambiguities.
Second, when given only the image with the target object blacked out ($Image$-$Only$, $Context$), the LLM achieves 24.93\% accuracy, demonstrating that even without textual descriptions, it can leverage visual context to some degree. However, this performance is still significantly lower than the ResNet Context-Only baseline (38.45\%), suggesting that the structured inductive biases of CNNs are better suited for this particular visual reasoning task than the general-purpose architecture of the LLM.
Third, the LLM still falls far short of our GSCG model, confirming that simply having access to the pixels is not enough; a structured representation enabling explicit reasoning is superior.
Fourth, in both text-only and multimodal settings, the LLM's best performance is not on the $full\_model$ prompt, but on an ablated one ($no\_colors$ for text-only, $no\_geometry$ for multimodal). This suggests the LLM is overwhelmed by the detail in the $full\_model$ description and performs better with a more concise, abstract prompt. This shows that for complex reasoning, the structure of the data is as important as the data itself. A neural network operating on graphs designed to handle such structure is inherently better suited to the task than an LLM that must parse a linearized, textual representation of that same structure.

\subsection{Limitations}
A key limitation in our evaluation is the inability to test against an "oracle" graph built from ground-truth data. This is because the datasets for semantic/instance segmentation (COCO) and material segmentation (DMS) are disjoint. Consequently, no single image in our dataset has ground-truth labels for both object identity and material composition, making it impossible to construct a perfect ground-truth GSCG. Such an experiment would have been valuable for isolating the performance of the $GraphObjectClassifier$ from errors introduced by the upstream perception models. Future work could explore newly created datasets that provide unified annotations for all these modalities.

\section{Human vs. AI: A Contextual Riddle Game}

To further assess the capabilities of our context-aware model, we designed a game where human players competed against our GSCG-based system. The game involved guessing a target object in a real-world image, often from outside the COCO dataset's distribution, based on a context-based riddle.

\paragraph{Game Setup.}
An image was processed through our GSCG pipeline. A target object was selected, and a riddle was generated by an LLM (e.g. Gemma 3 12B \cite{team2025gemma}) using the same structured, natural language description of the object's context that was provided to the Llama 4 Scout model in our main experiments. This riddle described the object's physical properties and its relationships with neighboring objects. Human players were presented with this riddle and a list of five possible choices for the object's identity. This list included the top five predictions from our $full\_model$; if the correct answer was not among them, it replaced one of the incorrect options.

\paragraph{Performance.}
Over a total of 60 rounds, human players achieved an accuracy of \textbf{68.33\%}. This result is noteworthy, especially considering the challenging nature of the task, which involved out-of-distribution images and objects. In the same set of rounds, the AI opponent, using the $full\_model$'s top-1 prediction, achieved an accuracy of \textbf{46.67\%}.

\paragraph{Discussion.}
The results of this experiment offer a nuanced view of the model's capabilities. It is crucial to recognize the asymmetry in the tasks: the AI performs a 1-of-80 classification to achieve its 46.67\% accuracy, while humans perform a 1-of-5 multiple-choice selection where the correct answer is always present. The human accuracy of 68.33\% is achieved with two significant aids: the multiple-choice format and an LLM-generated riddle that translates the GSCG's structured data into intuitive, natural language.

The AI's 46.67\% top-1 accuracy is therefore a more direct measure of its raw predictive power on this challenging, out-of-distribution dataset. The fact that this level of performance is achieved underscores the robustness of the GSCG-based approach. The experiment highlights that while the model's structured reasoning is powerful, there is a distinct advantage in the human ability to process descriptive, natural-language context. This suggests that a key area for future work is not just improving the underlying graph representation, but also developing methods to better translate this structured knowledge into more intuitive formats, enhancing human-AI collaboration.

\begin{table}[t]
\centering
\resizebox{\columnwidth}{!}{%
\begin{tabular}{@{}lc@{}}
\toprule
\textbf{Model} & \textbf{Accuracy (\%)} \\
\midrule
\multicolumn{2}{l}{\textit{Our GSCG-based Models (Full Context)}} \\
\textbf{GSCG Classifier (Full Model)} & \textbf{73.43 ± 0.32} \\
\midrule
\multicolumn{2}{l}{\textit{Baseline Models}} \\
ResNet (Combined 6-channel) & 53.52 ± 0.44 \\
Llama 4 Scout (Multimodal, Zero-Shot) & 42.34 ± 0.37 \\
ResNet (Object-Only, no context) & 44.10 ± 0.42 \\
ResNet (Full Context-Only, no object) & 38.45 ± 0.40 \\
Llama 4 Scout (Text-Only, Few-Shot) & 28.40 ± 0.32 \\
Llama 4 Scout (Image-Only, Context) & 24.93 ± 0.35 \\
Llama 4 Scout (Text-Only, Zero-Shot) & 23.16 ± 0.31 \\
\midrule
\multicolumn{2}{l}{\textit{Our GSCG-based Models (Ablated)}} \\
GSCG Classifier (no neighbors/global context) & 56.66 ± 0.37 \\
GSCG Classifier (minimal model, no object information, \\ 
no geometry, only with labels of immediate neighbors) & 38.41 ± 0.36 \\
\bottomrule
\end{tabular}%
}
\caption{High-level comparison of our GSCG-based classifier against strong baselines. Our full context-aware model significantly outperforms all baselines. Bootstrap confidence intervals (N=1000) are shown for our GSCG models. Even severely ablated versions of our model remain competitive, highlighting the value of our structured representation.}
\label{tab:main_results}
\end{table}

\begin{table*}[t]
\centering
\small
\begin{tabular}{@{}lcccc@{}}
\toprule
& \multicolumn{3}{c}{\textbf{Llama 4 Scout Performance (\%)}} & \textbf{GSCG Classifier (\%)} \\
\cmidrule(lr){2-4} \cmidrule(lr){5-5}
\textbf{Ablation Configuration} & \textbf{Zero-shot (Text)} & \textbf{Few-shot (Text)} & \textbf{Multimodal (Vision+Text)} & \textbf{(This Work)} \\
\midrule
full\_model & 18.26 ± 0.30 & 25.48 ± 0.30 & 37.31 ± 0.39 & \textbf{73.43 ± 0.32} \\
\midrule
no\_geometry & 22.28 ± 0.32 & 22.28 ± 0.30 & 42.34 ± 0.37 & \textbf{67.70 ± 0.33} \\
no\_colors & 21.04 ± 0.30 & 28.40 ± 0.32 & 38.55 ± 0.37 & \textbf{72.32 ± 0.32} \\
no\_colors\_no\_geometry & 23.16 ± 0.31 & 27.14 ± 0.33 & 41.95 ± 0.35 & \textbf{65.87 ± 0.34} \\
no\_materials & 8.82 ± 0.20 & 9.09 ± 0.21 & 29.65 ± 0.34 & \textbf{61.84 ± 0.35} \\
no\_materials\_no\_geometry & 10.13 ± 0.23 & 7.07 ± 0.18 & 33.94 ± 0.36 & \textbf{48.07 ± 0.36} \\
\midrule
no\_neighbors & 17.29 ± 0.28 & 26.73 ± 0.34 & 36.27 ± 0.35 & \textbf{71.33 ± 0.33} \\
no\_neighbors\_no\_geometry & 21.74 ± 0.31 & 24.56 ± 0.32 & 41.28 ± 0.35 & \textbf{63.82 ± 0.35} \\
no\_extended\_context & 16.59 ± 0.27 & 25.43 ± 0.31 & 34.75 ± 0.36 & \textbf{68.55 ± 0.34} \\
no\_extended\_context\_no\_materials & 7.30 ± 0.19 & 9.83 ± 0.23 & 25.62 ± 0.32 & \textbf{53.08 ± 0.36} \\
\midrule
no\_neighbors\_no\_extended\_context & 13.91 ± 0.26 & 25.95 ± 0.30 & 34.87 ± 0.34 & \textbf{56.66 ± 0.37} \\
minimal\_model & 8.34 ± 0.20 & 5.91 ± 0.18 & 30.91 ± 0.35 & \textbf{38.41 ± 0.36} \\
\bottomrule
\end{tabular}
\caption{Comprehensive performance comparison across different models and modalities. Bootstrap confidence intervals (N=1000) are shown for all models. Our GSCG classifier dramatically outperforms the LLM in all comparable settings. Note that for the LLM, simplifying the textual prompt (e.g., $no\_geometry$ or $no\_colors$) can paradoxically improve performance, suggesting it struggles to parse overly complex, structured descriptions, a task for which our graph-based model is purpose-built.}
\label{tab:llm_comprehensive}
\end{table*}

\section{Towards Human-Level Understanding of Objects in Context}
\label{sec:human_context}

Our work is inspired by the principles of active inference, a framework that states the brain constantly tries to integrate object properties with contextual information to form a cohesive understanding of its environment \cite{friston2010free, zhao2023survey}. Perception is then the process of updating the brain's internal model to best explain the observed sensory data. A key aspect of this process is belief updating, where prior beliefs about the world are combined with new evidence to form a more accurate posterior belief.

This provides a compelling explanation for how humans develop such a profound understanding of objects in context. Our perception is not a passive registration of sensory input but an active, inferential process \cite{chang2023survey}. The brain constantly generates predictions about the causes of its sensory signals, from low-level features like edges and colors to high-level concepts like object identities, their interrelations and relative positions. The discrepancy between these predictions and the actual sensory input—the prediction error—drives learning and belief updating.


Our Geo-Semantic Scene Graphs (GSCGs) can be seen as a formal, computational instantiation of the brain's generative model for scenes. The nodes in the graph represent hypotheses about the objects (the "what"), while the edges and their attributes represent the predicted relationships between them (the "where" and "why"). By explicitly encoding geometry, materials, and semantic relationships, the GSCG provides a rich, structured prior that enables more efficient and accurate inference, similar to knowledge-enhanced models \cite{chen2019knowledge, yang2023harnessing}.

When our $GraphObjectClassifier$ predicts the identity of a masked object, it is, in essence, performing a form of belief updating analogous to this perceptual inference process. It uses the contextual information encoded in the graph—the relationships with surrounding objects—as evidence to update its "belief" about the identity of the missing object. The success of our $full\_model$ demonstrates that a richer generative model—one that includes detailed information about geometry, materials, and both local and extended context—leads to more accurate inference. This aligns with the idea that perception relies on explaining away sensory input using a sophisticated internal model of the world, where context is crucial for resolving ambiguity.

\section{Conclusion and Future Work}

In this paper, we demonstrated the profound impact of context on object recognition. By moving beyond pixel-based classification and towards a structured, relational understanding of scenes, we achieve a new level of performance and robustness. Our main contribution is the Geo-Semantic Contextual Graph (GSCG), a rich, queryable world model built from a single 2D image, and a specialized graph classifier that leverages it. Our experiments unequivocally show that this explicit, structured approach to context dramatically outperforms both powerful but context-agnostic CNNs and state-of-the-art LLMs attempting to reason over unstructured visual and textual data.

This work opens up several exciting avenues for future research. The most promising is \textbf{novel class detection and description}. When encountering an unknown object, its GSCG node would lack a semantic label but would still be populated with rich contextual information (e.g., "it's a small metal object, on a desk, touching a keyboard"). A system could leverage this context to infer the object's likely function or even propose a descriptive name (e.g., "desk peripheral"), effectively learning about new objects from their environment. This moves beyond simple classification towards true scene understanding and knowledge acquisition, a critical step towards more general and adaptable AI.

\paragraph{Acknowledgements.} 
This work was supported in part by NeoArt Connect NAC 2025 Scholarship and by projects “Romanian Hub for Artificial Intelligence - HRIA”, Smart Growth, Digitization and Financial Instruments Program, 2021-2027 (MySMIS no. 334906) and "European Lighthouse of AI for Sustainability - ELIAS", Horizon Europe program (Grant No. 101120237).  

The authors would also like to thank Prof. Serena Stan from Instituto Cervantes (Bucharest) for her inspiring ideas and work, as team member, on the development of the visual riddle Human-AI game (Section 5) - an educational tool entitled "GuessWhat - Riddle Eye with AI", which was the winner (1st place) of the NeoArt Connect NAC 2025 Scholarship Program.

\appendix
\section{Appendix}
\subsection{GraphObjectClassifier Architecture Details}
The $GraphObjectClassifier$ is a modular classifier designed to handle the complex, attributed nature of our GSCG nodes. It is composed of several embedders that process different types of features, which are then aggregated for a final classification. The $base\_embed\_dim$ of 32 serves as a fundamental building block dimension for the embeddings of different features (label, geometry, material). Each feature type is initially projected into a latent space of this dimension.

The core components are:
\begin{itemize}
    \item \textbf{ColorEmbedder}: This module embeds RGB colors. It first converts RGB values to the CIELAB color space. A small MLP with two hidden layers (\(Linear(3) \rightarrow ReLU \rightarrow Linear(64) \rightarrow ReLU \rightarrow Linear(32)\)) then maps the 3 CIELAB coordinates to a 32-dimensional embedding. The final color representation for a material part is a weighted average of these embeddings, based on the percentage of each color on the material's surface.
    
    \item \textbf{MaterialEmbedder}: This module embeds material information. It combines a semantic material label embedding with a color embedding. It takes the one-hot encoded material label (from a vocabulary of the materials from the DMS dataset) and passes it through a linear layer to get a 32-dim vector. It uses the $ColorEmbedder$ to get a 32-dim vector for the material's dominant colors. These two vectors are concatenated and passed through a final MLP (\(Linear(64) \rightarrow ReLU \rightarrow Linear(64)\)) to produce the material embedding. The final representation for an object is a weighted average of its constituent material embeddings.

    \item \textbf{ObjectEmbedder}: This module creates a unified embedding for a single object. It combines embeddings for the object's semantic label (if available), its geometric properties, and its material composition.
    \begin{itemize}
        \item \textit{Label}: A one-hot encoded class label (from the COCO ontology) is passed through a linear layer to produce a 32-dim embedding. For the target object, this is a zero vector.
        \item \textit{Geometry}: A 15-dimensional vector (3 for size, 9 for flattened orientation matrix, 3 for center coordinates) is processed by a two-layer MLP (\(Linear(15) \rightarrow ReLU \rightarrow Linear(32)\)) to get a 32-dim embedding.
        \item \textit{Materials}: The $MaterialEmbedder$ provides a 64-dim embedding for the object's material composition.
    \end{itemize}
    These embeddings are concatenated and passed through a final two-layer object MLP to produce the final object feature vector.

    \item \textbf{GraphObjectClassifier}: This is the main classification model.
    \begin{itemize}
        \item It uses the $ObjectEmbedder$ to get feature vectors for the target node (without its label) and all its neighbors (with their labels).
        \item For each neighbor, it creates a combined representation by concatenating the neighbor's object embedding, a 16-dim embedding for the edge type ($touch$, $near$, $mixed$), and the scalar edge weight.
        \item This set of neighbor representations is processed by a standard $nn.MultiheadAttention$ layer with 8 heads to produce an aggregated local context vector.
        \item A global context vector is computed from a histogram of all object labels in the scene, which is passed through a two-layer MLP (\(Linear(num\_classes) \rightarrow ReLU \rightarrow Linear(64)\)).
        \item Finally, it concatenates the target's own embedding, the aggregated local context vector, and the global context vector. This is fed into a final classifier MLP with two hidden layers and Dropout (0.3) to predict the target object's class.
    \end{itemize}
    
    \item \textbf{NeighborlessObjectClassifier}: An ablation model that omits the local neighborhood aggregation step. It only uses the target object's own features (from the $ObjectEmbedder$) and the global context vector for classification.
\end{itemize}

\subsection{Hyperparameters}
The key hyperparameters for the $GraphObjectClassifier$ and its modules are:
\begin{itemize}
    \item \textbf{Training Hyperparameters}:
    \begin{itemize}
        \item Optimizer: AdamW
        \item Learning Rate: 1e-4
        \item Weight Decay: 1e-4
        \item Batch Size: 512
        \item Number of Epochs: 20 (with early stopping on validation F1-score, patience 10)
    \end{itemize}
    \item \textbf{Architectural Hyperparameters}:
    \begin{itemize}
        \item Base Embedding Dimension ($base\_embed\_dim$): 32
        \item Geometric Embedding Dimension: 32
        \item Color Embedding Dimension: 32
        \item Material Label Embedding Dimension: 32
        \item Total Material Embedding Dimension: 64
        \item Edge Type Embedding Dimension: 16
        \item Global Context Embedding Dimension: 64
        \item Attention Heads: 8
        \item Classifier Hidden Dimension: 256
        \item Classifier Dropout Rate: 0.3
    \end{itemize}
\end{itemize}

\subsection{Color Extraction Details}
Dominant colors are extracted using a k-means clustering approach on the CIELAB color space of an object's pixels.
\begin{itemize}
    \item K-means is initialized with $k$-$means{++}$ and a random seed of 42.
    \item After initial clustering (k=20), clusters are iteratively merged if their distance in CIELAB space is below a threshold of 15.0 (using CIEDE2000 distance).
    \item The final cluster centers are mapped to CSS4 color palette names, by finding the nearest color in this palette.
\end{itemize}

\subsection{Outlier Filtering}
The Z-score for outlier filtering on the point clouds is computed independently for each of the x, y, and z axes. A point is removed if its Z-score along any axis exceeds the threshold of 2.5.

{
    \small
    \bibliographystyle{ieeenat_fullname}
    \bibliography{main}
}

\end{document}